\documentclass{article} % For LaTeX2e
\usepackage{iclr2023_conference,times}
\usepackage{amsmath,amsfonts,bm}

\def\eqref#1{equation~\ref{#1}}
\def\1{\bm{1}}

\DeclareMathAlphabet{\mathsfit}{\encodingdefault}{\sfdefault}{m}{sl}
\SetMathAlphabet{\mathsfit}{bold}{\encodingdefault}{\sfdefault}{bx}{n}

\usepackage{hyperref}
\usepackage{url}
\usepackage{graphicx}
\usepackage{booktabs}
\usepackage{adjustbox}
\usepackage{multirow}

\usepackage{subcaption}
\newcommand{\lang}[1]{\textsc{#1}}

\title{Leveraging Auxiliary Domain Parallel Data in Intermediate Task Fine-tuning for Low-resource Translation}

\author{Shravan Nayak\textsuperscript{$\ddagger$}\thanks{Correspondence to: Shravan Nayak \texttt{<shravannayak.p@gmail.com>}} , Surangika Ranathunga\textsuperscript{$\dagger$}, Sarubi Thillainathan\textsuperscript{$\dagger$}, Rikki Hung\textsuperscript{$\mathsection$}, \\  \textbf{Anthony Rinaldi\textsuperscript{$\mathsection$}, Yining Wang\textsuperscript{$\mathsection$}, Jonah Mackey\textsuperscript{$\mathsection$}, Andrew Ho\textsuperscript{$\mathsection$}, En-Shiun Annie Lee\textsuperscript{$\mathsection$}} \\
\textsuperscript{$\ddagger$}Microsoft, \textsuperscript{$\dagger$}University of Moratuwa, \textsuperscript{$\mathsection$}University of Toronto
}

\iclrfinalcopy % Uncomment for camera-ready version, but NOT for submission.
\begin{document}

\maketitle

\begin{abstract}
NMT systems trained on Pre-trained Multilingual Sequence-Sequence (PMSS) models flounder when sufficient amounts of parallel data is not available for fine-tuning. This specifically holds for languages missing/under-represented in these models. The problem gets aggravated when the data comes from different domains. In this paper, we show that intermediate-task fine-tuning (ITFT) of PMSS models is extremely beneficial for domain-specific NMT, especially when target domain data is limited/unavailable and the considered languages are missing or under-represented in the PMSS model. We  quantify the domain-specific results variations using a domain-divergence test, and show that ITFT can mitigate the impact of domain divergence to some extent.
\end{abstract}

\section{Introduction}
Pre-trained Multilingual Sequence-Sequence (PMSS) models such as mBART~\citep{tang2020multilingual} and mT5~\citep{xue2021mt5} have shown considerable promise over vanilla Transformer models for Neural Machine Translation (NMT). This promise persists to low-resource language translation ~\citep{thillainathan2021fine}, which remains a challenge despite the recent advances in the field~\citep{ranathunga2021neural}. 

In addition to the empirical analysis carried out during the introduction of these PMSS models~\citep{tang2020multilingual}, further empirical analysis for NMT was conducted by ~\citet{wang2022understanding, Liu_Winata_Cahyawijaya_Madotto_Lin_Fung_2021} and \citet{lee2022pre}. The latter two focused on low-resource language pairs and showed that results for languages unseen in the PMSS model are very low.~\citet{lee2022pre} attributed the results diversity to the amount of fine-tuning/pre-training data, noise in fine-tuning data, domain mismatch in training/testing data and language typology. ~\citet{Liu_Winata_Cahyawijaya_Madotto_Lin_Fung_2021} experimented with continuous pre-training (CPT) to include unseen languages into the model, but found that when the amount of parallel data used in the fine-tuning stage is very low, CPT is not effective, particularly for non-English-centric translations.

However, both \citet{lee2022pre} and~\citet{Liu_Winata_Cahyawijaya_Madotto_Lin_Fung_2021} considered only the case where the PMSS model is fine-tuned only once with a parallel dataset belonging to a particular domain. A look into the available corpora suggests that there are either noisy automatically created parallel corpora or manually curated small parallel corpora for hundreds of languages, across diverse domains~\citep{tiedemann-thottingal-2020-opus}.~\citet{bapna2022building} automatically mined bitext for over 1000 languages from the web.~\citet{artetxe2020call} also point to several initiatives aimed at creating parallel resources at scale. This means, that for a given language pair, there can be several parallel datasets, belonging to different domains. In fact,~\citet{artetxe2020call} argue that pure unsupervised NMT setup is not realistic given the availability of parallel data. 

One way to utilize parallel data coming from auxiliary domains in building domain-specific NMT systems is Intermediate Task Fine-tuning (ITFT). In ITFT, the PMSS model is first fine-tuned with data from an auxiliary domain, and later fine-tuned on the target domain. The existing ITFT research for NMT considered only the case where the NMT model is tested only for a domain included in fine-tuning~\citep{verma2022strategies, adelani2022few}. Moreover, none of this research analysed the impact of domain divergence in ITFT. The other alternative to exploit auxiliary domain parallel corpora is to improve the pre-training stage of the PMSS model~\citep{reid2021paradise}. However, they do not discuss the impact of the size and domain of the parallel data used during pre-training.

The objective of this research is to explore the benefit of ITFT for domain-specific low-resource language NMT (LRL-NMT). We experimented with several LRLs, where some are not in the selected PMSS model. We selected datasets from three  domains and investigated the impact of ITFT on in-domain (train and test with the same domain data)  and out-domain (train with a dataset of one domain and test with another)  translation. The latter refers to zero-shot translation for the considered domain. 

We show that ITFT is beneficial for domain-specific translation when the target domain data is unavailable/limited, specifically for languages missing/under-represented in the PMSS models. We also note a strong impact of divergence between the domains of final task and testing data. However, using a different domain for the ITFT dampens the impact of this domain divergence. Thus when multiple domain data is available for a language pair, consciously selecting data for ITFT is important. We also release a multi-way parallel bible dataset of 25k for the selected low-resource languages, which until now had less than 4k parallel sentences. 
\section{Related Work}
\label{related_work}
Before the PMSS era, researchers have experimented with Transfer Learning (TL) on vanilla Transformer~\citep{vaswani2017attention} models and recurrent models.  During TL, a low-resource translation task is trained on an NMT model, which has already been trained for a high-resource translation task~\citep{lakew2018transfer, dabre-etal-2019-exploiting, maimaiti2020enriching, imankulova-etal-2019-exploiting}. 

ITFT is based on the concept of TL, where a pre-trained language model is fine-tuned with an intermediate task, before the final task. ITFT has been extensively studied for encoder-based models such as BERT~\citep{phang2018sentence}.~\citet{takeshita2022x,maurya2021zmbart},  used ITFT on PMSS models for summarization and text generation, respectively. They investigated the use of auxiliary \textit{tasks} in ITFT. As for NMT, ~\citet{adelani2022few} explored two variants: (1) fine-tune with data from an auxiliary domain and further fine-tune with the target domain and (2) fine-tune by combining data from different domains and further fine-tune with the target domain. However, they tested only for the in-domain setup. ~\citet{verma2022strategies} also experimented with data from one domain as the intermediate task, and with a multilingual intermediate task. In the latter, data comes from multiple languages but for the same domain. However, none of this research had a detailed look at the impact of domain difference of datasets.

\section{Experimental Settings }

\begin{figure}[htp]
    \centering
    \includegraphics[scale=0.5]{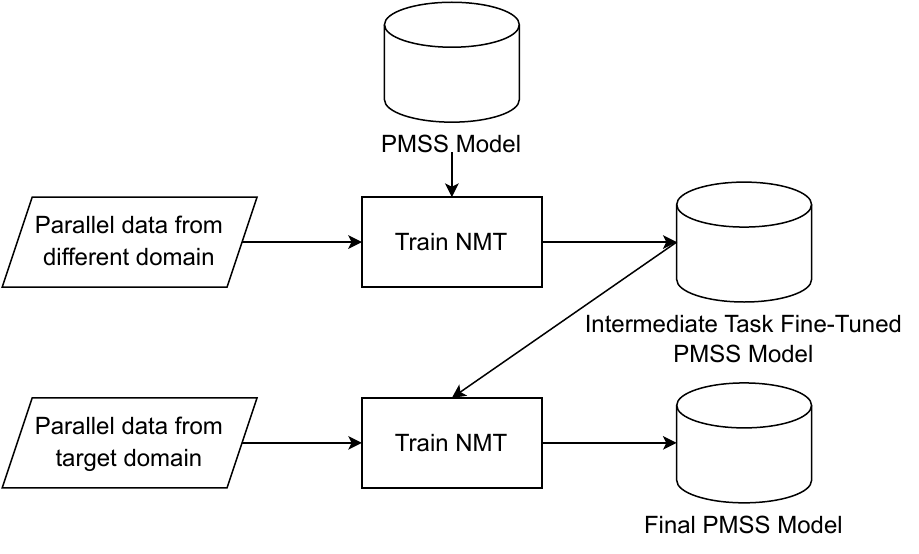}
    \caption{Overview of Multistage Fine-tuning.}
    \label{fig:multistage_ft_pt}
\end{figure}

\paragraph{ITFT:}  We fine-tune the PMSS model first with an out-domain  parallel dataset (also referred to as \textit{intermediate task}), followed by parallel data from another domain (also referred to as \textit{final task}; see Figure~\ref{fig:multistage_ft_pt}).  Testing is carried out with data from the second domain (in-domain) or a different domain (out-domain).
\paragraph{Languages:}
We select four low-resource languages (Gujarati, Kannada, Sinhala and Tamil ~\citep{joshi-etal-2020-state}) along with English and Hindi. All, except English use non-Latin scripts~\citep{pires-etal-2019-multilingual}. Table \ref{tab:lang-details} reports further details.

\begin{table}[t]
\centering
%\small
\small
\scalebox{0.8}{ %0.73
\begin{tabular}{lllrrr} 
\toprule
Language & Family      & Script     & Joshi & mBART &mT5  \\
    &      &    & class  & Tokens (M) & Tokens (M) \\\midrule
Hindi (\lang{\lang{hi}})       & IA  & Devanagari & 4 & 1715    & 24000   \\
Gujarati (\lang{\lang{gu}}) & IA & Gujarati & 1 & 140 & 800 \\
Kannada (\lang{\lang{kn}})       & Dr       & Kannada  & 1  & -- & 1100  \\
Sinhala (\lang{\lang{\lang{si}}})       & IA  & Sinhala  & 1  & 243 & 800     \\
Tamil (\lang{\lang{ta}})       & Dr  & Tamil   & 3   & 595 &3400       \\
\bottomrule
\end{tabular}
}
\caption{Languages (IA- Indo Aryan, Dr - Dravidian)}
\label{tab:lang-details}
\vspace{-0.2cm}
\end{table}
\begin{table}[t]
\centering
\begin{adjustbox}{max width=\linewidth}
\footnotesize
\begin{tabular}{@{} l @{\qquad} l @{\qquad} l @{\qquad} l @{\qquad} l @{}} \toprule
Dataset               & Domain & Languages & Train Size & Test Size \\ \midrule
\textsc{Flores}-101                & Open       &  \lang{hi, gu, kn, ta}          & test only & 1k \\
\textsc{Flores}v1                & Open       & \lang{si}   & test only   & 1k\\
\addlinespace
CCAligned (CC)            & Open       & \lang{all}   & 100k         & 1k \\
\addlinespace
Government (Gvt)            & Administrative       & \lang{\lang{si}}, \lang{ta}          & 50k    & 1k \\
PMIndia (PMI)              & News       &  \lang{hi}           & 50k   & 1k\\
              &       &  \lang{gu}, \lang{kn}          & 25k  & 1k \\
\addlinespace
Web-scrap Bible                 & Religious       &  \lang{all}         & 25k    & 1k \\
\bottomrule
\end{tabular}
\end{adjustbox}
\caption{Dataset statistics}
\label{tab:data-details}
\vspace{-5mm}
\end{table}

\paragraph{Dataset:} We use a mix of both open-domain and domain-specific corpora to train and test our models. The domain-specific corpora differ across the family of languages. Dataset summary details are given in Table~\ref{tab:data-details}. More details about the datasets and our new dataset can be found in Appendix~\ref{app:datasets}. Note that PMI and Government (Gvt) corpora are mutually exclusive for the datasets we considered\footnote{Although Tamil data is available in the PMI corpus we do not use this for our experiments.}. Therefore, when describing results (Section \ref{results}), we use PMI/Gvt to denote that we use one of these corpora for the considered experiment.

\paragraph{PMSS Models:}
Past work reported mixed results on the comparative performance of the two commonly used PMSS models, mBART and mT5~\citep{lee2022pre, liu2021continual}. Thus we considered both these models for initial experiments. Comprehensive training details are given in Appendix \ref{app:hyperparams}.
\section{Results and Discussion} \label{results}
We use SentencePiece BLEU (spBLEU)~\citep{flores101} as the evaluation metric. Reasons for selecting this metric are discussed in Appendix~\ref{appendix:c}. For all our experiments, we obtain results for \lang{xx}-\lang{en}, as well as \lang{en}-\lang{xx} tasks. Observations discussed in this section hold for both translation directions, unless specifically mentioned. We carry out both \textit{out-domain testing} as well as \textit{in-domain testing}. Test set specifications are as indicated in Table~\ref{tab:data-details}.

\subsection{Baseline Results}
We fine-tune mBART and mT5 with each of the training sets, and evaluate with the test set. As per Figure~\ref{fig:mbart_vs_mt5} in Appendix~\ref{appendix:d}, mBART outperforms mT5 across domains and dataset sizes (except for Kannada, which is missing in mBART), for both in-domain and out-domain testing, and confirm~\citet{lee2022pre, liu2021continual}'s observations.
Thus we selected mBART for further experiments.

\begin{figure}
    \includegraphics[scale=0.22]{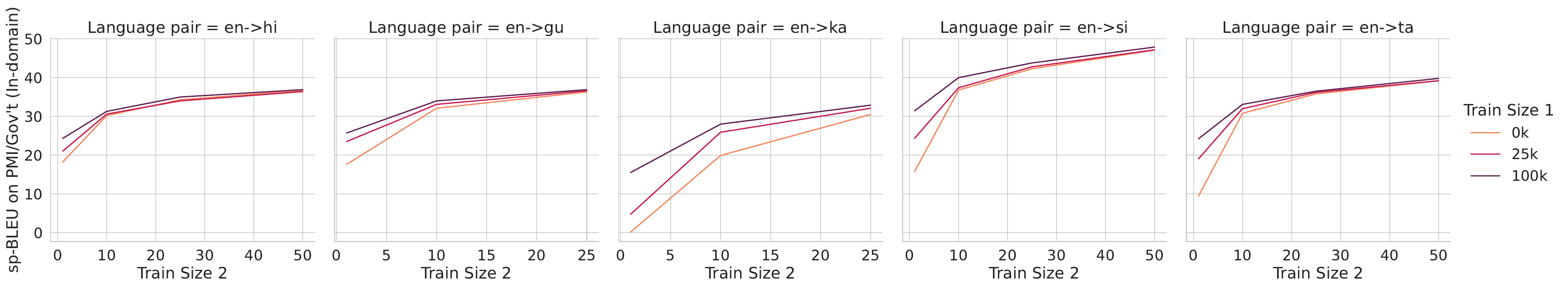}
     \caption{In-domain testing with ITFT. Intermediate task - CC, final task - PMI and test - PMI. X-axis shows the final task size. Colours show the intermediate task size (size 0 = baseline). Train Size 1 corresponds to size of CC data set and similarly Train Size 2 for PMI/Gvt. }
     \label{fig:intermediate_size_tiny}
\end{figure}

Our mBART results in Table \ref{tab:result-baseline} in Appendix~\ref{appendix:d} echo ~\citet{lee2022pre}'s observations: For the in-domain cases, NMT models built on mBART produce very low results for Kannada, when the parallel data size is less than 10k. However, with  25k parallel sentences, even for Kannada, the model reports very strong results. This strong result confirms the data efficiency of the models trained on mBART. However,  out-domain results are beyond usable levels for most of the cases. We analyse this observation with respect to domain divergence in Section~\ref{sec_divergence}.

\subsection{Effectiveness of ITFT}

We vary the size and domain of the intermediate task, as well as the size of the final task. 
\begin{figure}
     \centering
     \begin{subfigure}[b]{ \textwidth}
         \centering
         \includegraphics[scale=0.22]{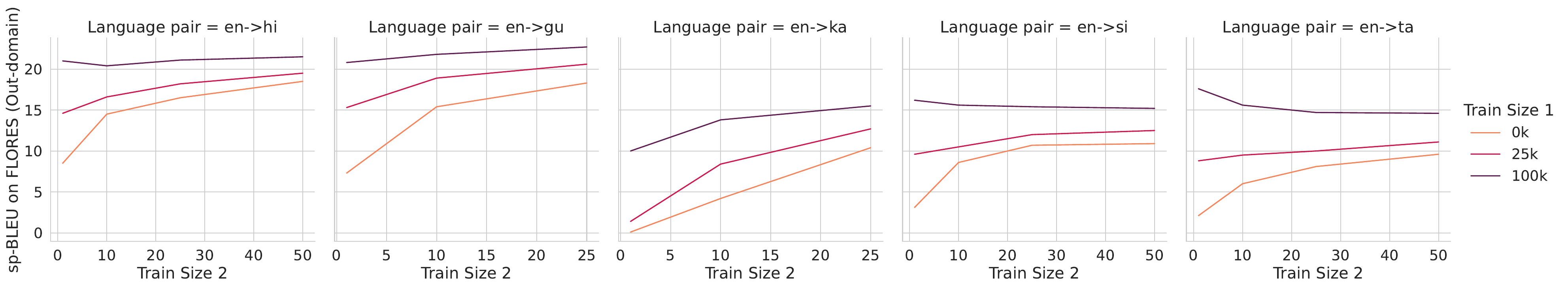}
        \caption{Intermediate Task Fine-Tuning with Intermediate task - CC, final task PMI. The spBLEU scores corresponds to scores on FLORES test set.} %Train 1 CC - Train 2 PMI
     \end{subfigure}
     \begin{subfigure}[b]{ \textwidth}
         \centering
         \includegraphics[scale=0.22]{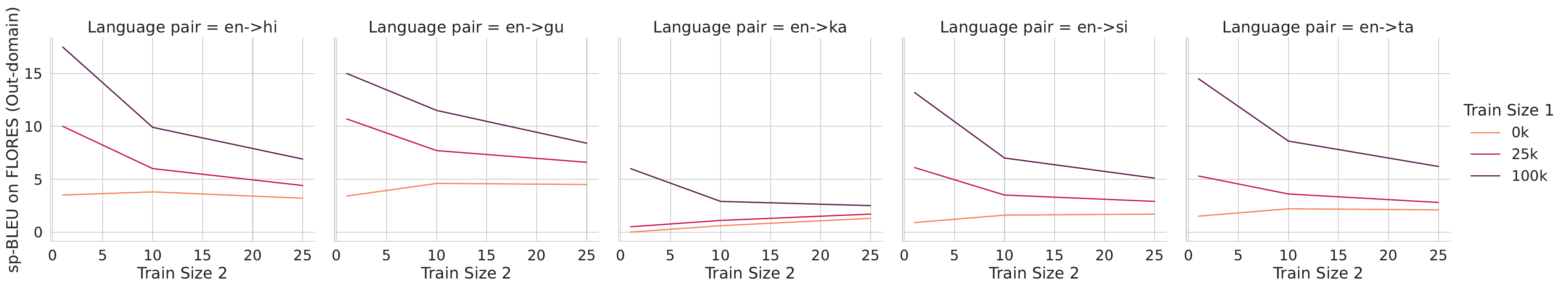}
        \caption{Intermediate Task Fine-Tuning with Intermediate task - CC, final task Bible. The spBLEU scores corresponds to scores on FLORES test set.} %Train 1 CC - Train 2 bib
     \end{subfigure}
     \caption{Out-domain testing with ITFT. intermediate task - CC,  test- FLORES, (a) final task - PMI, (b) final task - Bible. X-axis shows the final task size. Colours show the intermediate task size (size 0 = baseline). Train Size 1 corresponds to size of the intermediate task data set and similarly Train Size 2 for final task.}
     \label{fig:intermediate_size_tiny_out}
\end{figure}
\paragraph{In-domain translation:} Here, the domain of the final task is similar to that of the testing data. We considered PMI/Gov't and Bible domains for the final task, since they have data for both training and testing. Figure~\ref{fig:intermediate_size_tiny} shows the result when the intermediate task is CC and final task  is PMI. The complete result is reported in Figure ~\ref{fig:intermediate_size} in Appendix~\ref{appendix:d}. 

We see that ITFT with a different domain outperforms the baseline when the final task has less than 10k data (except for the 1k size of intermediate task for Ta and Si, where results are on par with the baseline). More training data for the intermediate task always helps in this setup. However, when final task datset size increases, the impact of intermediate task saturates and finally converges to the baseline result. The language-wise observation is worth noting - the impact of ITFT is the highest for Kannada. The gains over the baseline are visible even when the final task dataset size reaches 50k. However, gain of ITFT is the least for Hindi, which is relatively highly represented in mBART.

\paragraph{Out-domain translation:} In Figure~\ref{fig:intermediate_size_tiny_out} we show the results for the cases where the intermediate task is CC, the final task is PMI/Gov't and Bible, respectively, and the testing task is FLORES. Results for all the domain combinations are shown in Figure~\ref{fig:intermediate_size} in Appendix~\ref{appendix:d}. 

Similar to the in-domain setup, we see that ITFT is very beneficial when the final task has less than 10k parallel data, and having more data for the intermediate task is always beneficial. However, when the intermediate task has only 1k parallel sentences, its effect is on par with that of the baseline for some cases (e.g.~Kannada: intermediate - Bible, final - PMI). Different to in-domain translation, ITFT noticeably outperforms the baseline even for Hindi. However, we see that when Bible is used as the final task, the baseline shows very small gains as the final task size is increased. The gain slowly diminishes as the final task size is increased and finally tends to converge with the baseline result.

\subsection{Impact of Domain Divergence}
\label{sec_divergence}%@Yining
~\citet{lee2022pre} observed an impact of domain divergence in their results (i.e.~our baseline), however did not provide any quantitative analysis. 
In contrast, we systematically quantify the impact of domain difference. 

We measured the similarity between the two domains using the Jenson-Shannon (JS) divergence as defined in ~\citet{plank-van-noord-2011-effective}, which is a modification of the Kullback-Leibler (KL) divergence (See Appendix~\ref{app:js} for details of the calculation). Divergence between the training and test sets we used is given in Table~\ref{tab:result-domaindiv}.

To begin with, correlation between baseline results (Table~\ref{tab:result-baseline}) and the domain divergence values (Table~\ref{tab:result-domaindiv}) is $R^2 = 0.95$ -  a very strong correlation. For example, when FLORES is the test set, fine-tuning with PMI/Gov't gives much higher results than when using Bible, which is explained by the higher divergence between FLORES and Bible. 

In order to quantify the interplay between size and domain of intermediate task fine-tuning, we calculated the correlation values as shown in Table~\ref{tab:correlation}. When the size of the intermediate task is 25k, the divergence between the domain of the final task and the domain of the test task shows a lesser correlation to that of the baseline. This means ITFT slightly dampens the effect of domain divergence of the final task, although the domain of the intermediate task itself does not have a noticeable impact on the result. When the intermediate task size is 1k, it makes no impact on the divergence between the final and test task domains. This could be the reason why the 1k ITFT result lies closer to the baseline result. We also note that when the final task size is 1k, the correlation is not well-defined, despite the size of intermediate task.

\begin{table}[]
\parbox{.45\linewidth}{
\centering
\resizebox{6.5cm}{!}{
\begin{tabular}{lllll}
\toprule
Dataset     & Gvt test & FLORES test & Bib test & PMI test \\ \midrule
Gvt train & 0.10       & 0.40        & 0.58     & -       \\
CC train    & 0.40       & 0.41        & 0.52     & 0.44     \\
Bib train   & 0.56       & 0.47        & 0.10     & 0.53     \\
PMI train   & -          & 0.33        & 0.52     & 0.15     \\ \bottomrule
\end{tabular}
}
\caption{JS divergence between train and test domains}
\label{tab:result-domaindiv}
}
\hfill
\parbox{.45\linewidth}{
\vspace{0.38cm}
\centering
\resizebox{6.5cm}{!}{
\begin{tabular}{lll}
\toprule
Intermediate-final task size     & Intermediate task & Final task \\ \midrule
1k - 1k & 0.07       & 0.16  \\
25k - 25k    & 0.1       &  0.87   \\
1k - 25k  & 0.49      &  0.95   \\
25k - 1k  &  0.25     &  0.04    \\ \bottomrule
\end{tabular}
}
\caption{Correlation between the out-domain result and the divergence between train/test domains}
\label{tab:correlation}
}
\end{table}

\section{Conclusion}
We analysed the possibility of using auxiliary parallel data in an ITFT setup to improve the PMSS model performance for low-resource domain-specific NMT. Our results show that ITFT is beneficial when limited data is available for a particular domain. This observation strongly holds for languages missing or under-represented in the PMSS model. We also show that the domain differences play a major role in out-domain translation, however ITFT can mitigate this impact to a certain extent. In future, we expect to combine parallel data from different domains in a multi-task setup to further mitigate the impact domain divergence.

\bibliography{anthology, iclr2023_conference}
\bibliographystyle{iclr2023_conference}

\appendix
\section{Experiments}
\label{app:appendixA}

\subsection{Datasets} \label{app:datasets}

\paragraph{Bible corpus} Existing parallel corpora for Bible such as \citet{mccarthy2020johns}, although multiway parallel, have very little data for the languages we considered. Since we intend to perform a detailed analysis on dataset size, we curate a bible corpus for languages used in our experiments. We scrape Bible data the from web\footnote{Sinhala: https://www.wordproject.org/bibles/si/index.htm; and others: https://ebible.org/download.php} and then automatically align the sentences (at verse level). We were able to curate a multi-way parallel corpus of size 25k for 4 languages (\lang{kn}, \lang{gu}, \lang{hi}, \lang{ta}). Note that Sinhala was scraped from a different website, thus has different content\footnote{ The scripts to create the Bible dataset can be found at https://github.com/LRLNMT/BibleCorpus.}.

\paragraph{Common Crawl (CC)} CCAligned~\citep{el2020massive} corpus consists of parallel text that was automatically aligned using LASER sentence embeddings~\citep{schwenk-2018-filtering}.The dataset, although noisy~\citep{qualityAtAGlance}, has been used to develop highly multilingual machine translation models such as M2M100~\citep{fan2020englishcentric} and mBART multilingual MT~\citep{tang2020multilingual}.

\paragraph{PMIndia corpus (PMI)} PMIndia \citep{Barry2020PMIndia} is a parallel corpus for English and 13 other languages in India. It consists of news updates and excerpts of the Prime Minister's speeches extracted from the Prime Minister of India's website.

\paragraph{Government corpus (Gvt)} The government document corpus \citep{fernando2020data} is a multi-way parallel multilingual corpus for Sinhala, Tamil and English. It contains annual reports, committee reports, crawled content from government institutional websites, procurement documents, and acts from official Sri Lankan government documents.

\paragraph{FLORES} The \textsc{Flores}-101~\citep{goyal2021flores}  dataset is a multilingual, multi-way parallel corpus whose sentences are extracted from English Wikipedia and translated into 101 languages. It consists of data from a variety of topics and domains. We use \textsc{Flores}v1~\citep{guzman2019flores} for Sinhala since it is not present in \textsc{Flores}-101.

\subsection{Model Settings} \label{app:hyperparams}
We used HuggingFace libraries for our experiments.
All experiments were run with seed 222 and performed using a Nvidia Volta of 32 GPU RAM.

We train up to $3$ epochs with learning rate of $5 \cdot 10^{-5}$, dropout of $0.1$, maximum length of $200$ for the source and target, and a batch size of $10$ for training and evaluation. We use the implementations in the HuggingFace Transformers
library.

For kn language, we applied the related language fine-tuning strategy where we picked Telugu (te) for kn \citep{lee2022pre}.  

\subsection{Evaluation Metrics}
\label{appendix:c}
We use SentencePiece BLEU (spBLEU in short), introduced by \citet{flores101} as the evaluation metric for all our experiments. In this method, the BLEU scores are calculated for the text tokenized using sentence-piece subword model (which has been trained for all the 101 languages in \textsc{Flores}-101 dataset). The standardization of tokenizers allows research to make comparisons among each other. Further, \citet{flores101} also show that spBLEU functions similar to BLEU and also has strong correlation with the tokenzier-independent Chrf++ metric \citep{popovic-2017-chrf}. We use the official implementation provided in the sacreBLEU library\footnote{https://github.com/mjpost/sacreBLEU} \citep{post-2018-call} for evaluating all the experiments.

\subsection{Jenson-Shannon Divergence}
\label{app:js}
We calculate domain divergence using Jenson-Shannon (JS) divergence. JS divergence is calculated between two distributions $P$ and $Q$ using the formula: 
$$JSD(P||Q) = \frac{1}{2}KL(P||M) + \frac{1}{2}KL(Q||M)$$

$M$ represents an equally weighted sum of the two distributions (i.e., $M = \frac{1}{2}P + \frac{1}{2}Q$) and $KL(\cdot || \cdot)$ represents the Kullback–Leibler divergence.\\

The divergence is calculated for each (intermediate and fine-tuning) train and test set pair that were studied in the paper. The values are then averaged across all languages to obtain a similarity score that is independent of the target language. 
The train sets are composed of either 25k or 1k sentences and the test sets are composed entirely of 1k sentences. Each corpus is first tokenized using the \verb|NLTK| package,\footnote{https://www.nltk.org/} stripped of all stopwords, and transformed into a (discrete) frequency distribution over all word tokens. We also convert all times and numbers into the tokens \verb|<TIME>| and \verb|<NUMBER>| respectively. The frequency distributions of each train and test set are then compared using the formula above. JS divergence ranges from 0 to 1 with lower values indicating that the two distributions are more similar.
\clearpage

\section{Results}
\label{appendix:d}

\subsection{mBART vs mT5}
\begin{figure}[htp]
     \begin{subfigure}[b]{0.5\textwidth}
         \centering
           \includegraphics[width=.9\linewidth]{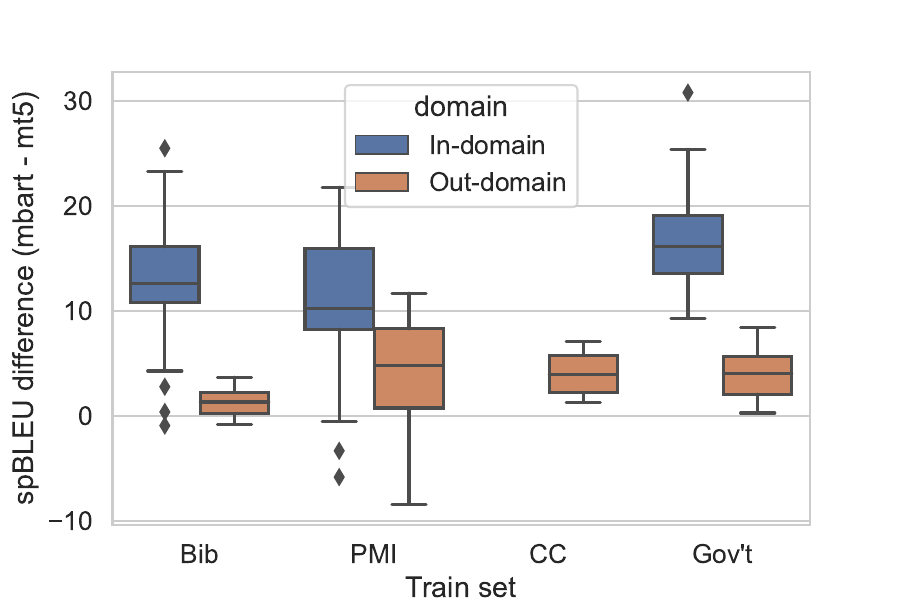}  
         \caption{Difference in performance by training set}
         \label{fig:traiing_set}
     \end{subfigure}
     \begin{subfigure}[b]{0.5\textwidth}
         \centering
           \includegraphics[width=.9\linewidth]{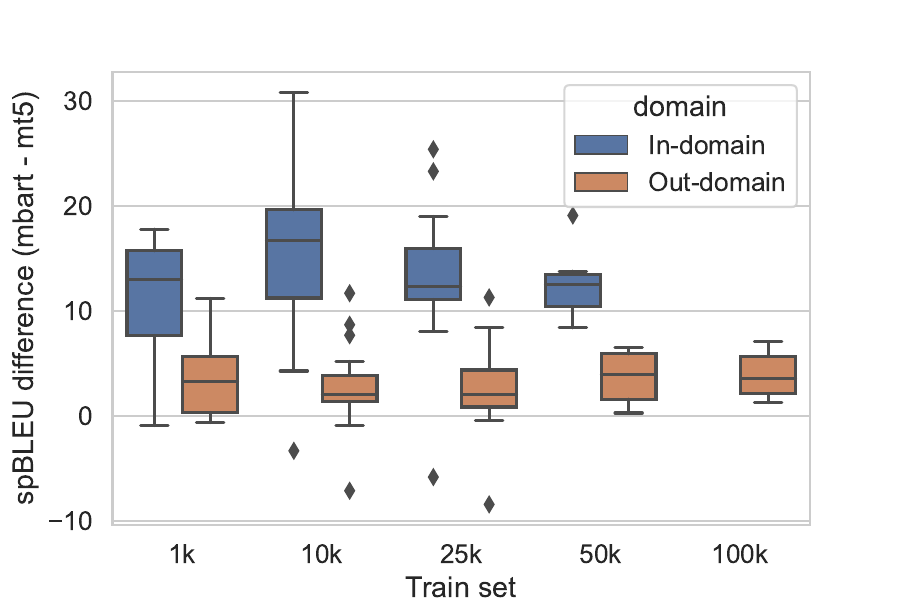}  
         \caption{Difference in performance by training set size}
         \label{fig:size}
     \end{subfigure}
        \caption{Comparative Analysis between mBART and mT5}
        \label{fig:mbart_vs_mt5}
\end{figure}

\subsection{Baseline}
\newlength\datgap
\setlength{\datgap}{2em}

\begin{table}[h]
\centering
\resizebox{\textwidth}{!}{
\begin{tabular}{@{}ll @{\hspace{\datgap}} lll @{\hspace{\datgap}} lll @{\hspace{\datgap}} lll @{\hspace{\datgap}} lll @{\hspace{\datgap}} lll}
\toprule
\multirow{3}{*}{Training}  & \multirow{3}{*}{Size} & \multicolumn{15}{c}{\lang{en} -\textgreater  \lang{xx}}                                                                                                    \\ \cmidrule(){3-17} 
                           &                       & \multicolumn{3}{c}{\lang{kn}} & \multicolumn{3}{c}{\lang{gu}} & \multicolumn{3}{c}{\lang{hi}} & \multicolumn{3}{c}{\lang{si}} & \multicolumn{3}{c}{\lang{ta}}                  \\ \cmidrule(r{\datgap}){3-5}  \cmidrule(r{\datgap}){6-8} \cmidrule(r{\datgap}){9-11} \cmidrule(r{\datgap}){12-14} \cmidrule(){15-17} 
                           &                       & FLORES  & Bib   & PMI  & FLORES  & Bib   & PMI  & FLORES  & Bib   & PMI  & FLORES  & Bib  & Gvt & FLORES                   & Bib  & Gvt \\ \hline
Zero shot                  & -                     & 0.1     & 0.0   & 0.1  & 0.3     & 0.0   & 0.1  & 0.3     & 0.0   & 0.4  & 0.2     & 0.0  & 1.2   & 0.5                      & 0.0  & 0.9   \\
\multirow{4}{*}{PMI/Gvt} & 1k                    & 0.1     & 0.0   & 0.1  & 7.3     & 2.1   & 17.6 & 8.5     & 1.7   & 18.2 & 3.1     & 0.4  & 15.7  & 2.1                      & 0.5  & 9.4   \\
                           & 10k                   & 4.2     & 0.5   & 19.9 & 15.4    & 3.7   & 32.1 & 14.5    & 2.9   & 30.2 & 8.6     & 1.1  & 36.8  & 6.0                      & 0.8  & 30.8  \\
                           & 25k                   & 10.4    & 1.1   & 30.5 & 18.3    & 4.4   & 36.3 & 16.5    & 3.1   & 34.3 & 10.7    & 1.1  & 42.3  & 8.1                      & 1.1  & 35.8  \\
                           & 50k                   & -       & -     & -    & -       & -     & -    & 18.5    & 3.4   & 36.7 & 10.9    & 1.0  & 47.1  & 9.6                      & 1.3  & 39.2  \\
\multirow{3}{*}{Bib}       & 1k                    & 0.0     & 3.6   & 0.0  & 3.4     & 10.0  & 3.8  & 3.5     & 13.3  & 4.6  & 0.9     & 10.9 & 0.9   & 1.5                      & 9.7  & 1.5   \\
                           & 10k                   & 0.6     & 17.3  & 0.4  & 4.6     & 23.2  & 3.5  & 3.8     & 26.5  & 3.1  & 1.6     & 30.9 & 1.2   & 2.2                      & 24.0 & 1.6   \\
                           & 25k                   & 1.3     & 24.9  & 0.5  & 4.5     & 27.3  & 3.2  & 3.2     & 30.5  & 2.4  & 1.7     & 36.1 & 1.0   & 2.1                      & 29.4 & 1.1   \\
\multirow{2}{*}{CC}        & 25k                   & 0.2     & 0.0   & 0.1  & 7.8     & 1.1   & 4.0  & 15.1    & 5.6   & 13.0 & 9.3     & 2.1  & 14.6  & 9.4                      & 5.7  & 7.8   \\
                           & 100k                  & 2.8     & 0.1   & 2.1  & 7.5     & 0.9   & 6.4  & 23.1    & 6.8   & 19.8 & 15.1    & 4.0  & 25.9  & 16.0                     & 8.4  & 15.2  \\ \bottomrule
                          
\multirow{3}{*}{Training}  & \multirow{3}{*}{Size} & \multicolumn{15}{c}{\lang{xx} -\textgreater \lang{en}}                                                                                                    \\ \cmidrule(){3-17}  
                           &                       & \multicolumn{3}{c}{\lang{kn}} & \multicolumn{3}{c}{\lang{gu}} & \multicolumn{3}{c}{\lang{hi}} & \multicolumn{3}{c}{\lang{si}} & \multicolumn{3}{c}{\lang{ta}}                  \\ \cmidrule(r{\datgap}){3-5}  \cmidrule(r{\datgap}){6-8} \cmidrule(r{\datgap}){9-11} \cmidrule(r{\datgap}){12-14} \cmidrule(){15-17}
                           &                       & FLORES  & Bib   & PMI  & FLORES  & Bib   & PMI  & FLORES  & Bib   & PMI  & FLORES  & Bib  & Gvt & FLORES                   & Bib  & Gvt \\ \midrule

Zero Shot                  & -                     & 0.1     & 0.0   & 0.0  & 0.3     & 0.0   & 0.2  & 0.3     & 0.0   & 0.5  & 0.2     & 1.2  & 0.0   & 0.3                      & 0.0  & 0.5   \\
\multirow{4}{*}{PMI/Gvt} & 1k                    & 0.4     & 0.0   & 2.4  & 9.6     & 2.8   & 17.2 & 12.7    & 4.4   & 20.5 & 6.2     & 1.9  & 16.2  & \multicolumn{1}{c}{5.4}  & 1.7  & 13.2  \\
                           & 10k                   & 1.5     & 0.5   & 12.3 & 13.7    & 2.8   & 17.2 & 16.6    & 5.2   & 32.0 & 7.3     & 2.7  & 30.7  & \multicolumn{1}{c}{8.4}  & 2.5  & 29.5  \\
                           & 25k                   & 2.3     & 0.7   & 14.9 & 13.8    & 4.0   & 31.8 & 18.4    & 5.0   & 36.5 & 9.0     & 2.5  & 37.1  & \multicolumn{1}{c}{8.2}  & 2.5  & 34.3  \\
                           & 50k                   & -       & -     & -    & -       & -     & -    & 19.3    & 5.4   & 38.6 & 9.2     & 2.6  & 41.0  & \multicolumn{1}{c}{10.0} & 3.0  & 39.4  \\
\multirow{3}{*}{Bib}       & 1k                    & 0.0     & 0.8   & 0.0  & 4.0     & 15.6  & 3.3  & 6.8     & 18.1  & 6.6  & 2.9     & 15.0 & 41.0  & \multicolumn{1}{c}{3.5}  & 13.2 & 2.3   \\
                           & 10k                   & 0.5     & 13.6  & 0.5  & 3.9     & 27.9  & 2.0  & 5.0     & 29.7  & 3.6  & 2.9     & 29.0 & 1.9   & \multicolumn{1}{c}{3.1}  & 27.2 & 1.4   \\
                           & 25k                   & 1.2     & 25.2  & 0.8  & 3.4     & 32.5  & 1.7  & 4.2     & 34.1  & 3.0  & 2.9     & 33.7 & 1.4   & \multicolumn{1}{c}{2.6}  & 32.7 & 1.1   \\
\multirow{2}{*}{CC}        & 25k                   & 0.4     & 0.0   & 0.3  & 5.0     & 0.8   & 5.3  & 15.0    & 7.3   & 11.9 & 8.4     & 3.5  & 12.0  & \multicolumn{1}{c}{8.2}  & 4.4  & 4.0   \\
                           & 100k                  & 0.5     & 0.0   & 0.8  & 5.0     & 1.2   & 4.8  & 19.6    & 8.6   & 15.4 & 12.7    & 6.6  & 20.2  & \multicolumn{1}{c}{13.8} & 8.5  & 9.8   \\ \bottomrule
\end{tabular}
}
\caption{Experimental results reported in spBLEU for the baseline. Zero shot corresponds to testing the mBART model without any finetuning. }
\label{tab:result-baseline}
\end{table}

\subsection{Intermediate Fine-tuning}

\begin{figure}
     \centering
     \begin{subfigure}[b]{ \textwidth}
         \centering
         \includegraphics[scale=0.22]{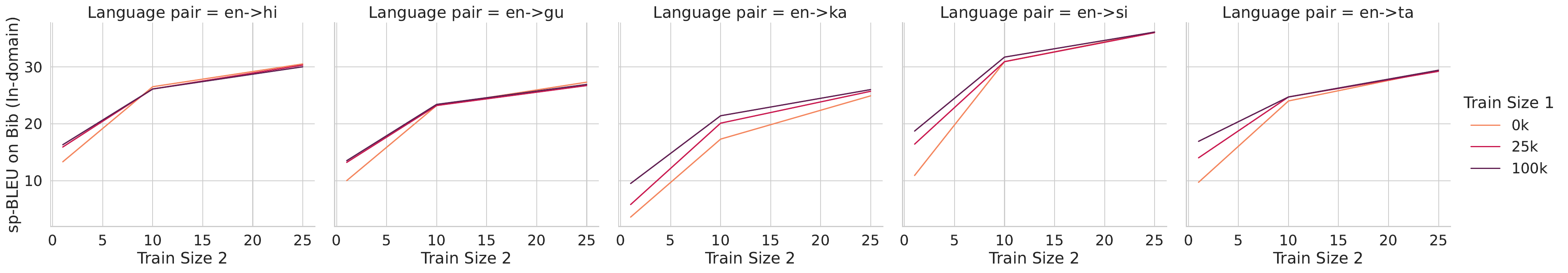}
        \caption{Intermediate Task Fine-Tuning with intermediate task - CC and final task - Bible. The spBLEU scores corresponds to scores on Bible test set.}
     \end{subfigure}
     \begin{subfigure}[b]{ \textwidth}
         \centering
         \includegraphics[scale=0.22]{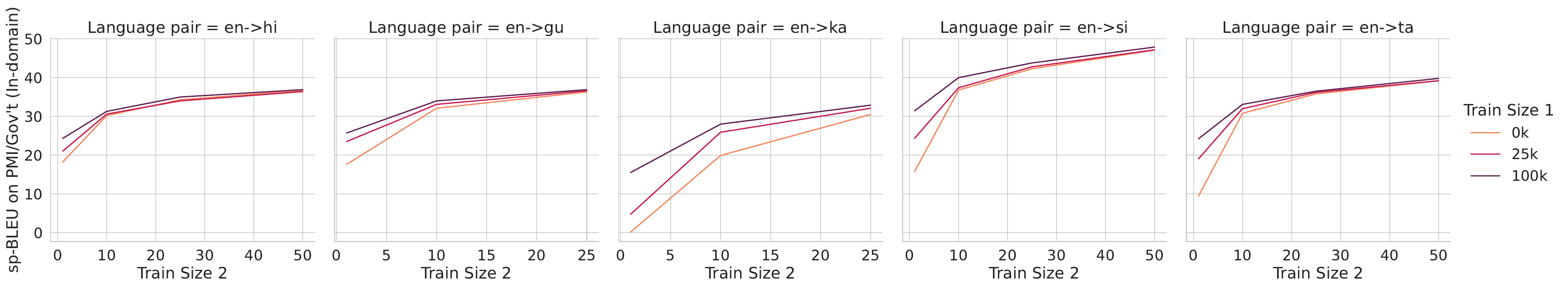}
        \caption{Intermediate Task Fine-Tuning with intermediate task - CC and final task - PMI/Gvt. The spBLEU scores corresponds to scores on PMI/Gvt test set.}
     \end{subfigure}
     \begin{subfigure}[b]{ \textwidth}
         \centering
         \includegraphics[scale=0.22]{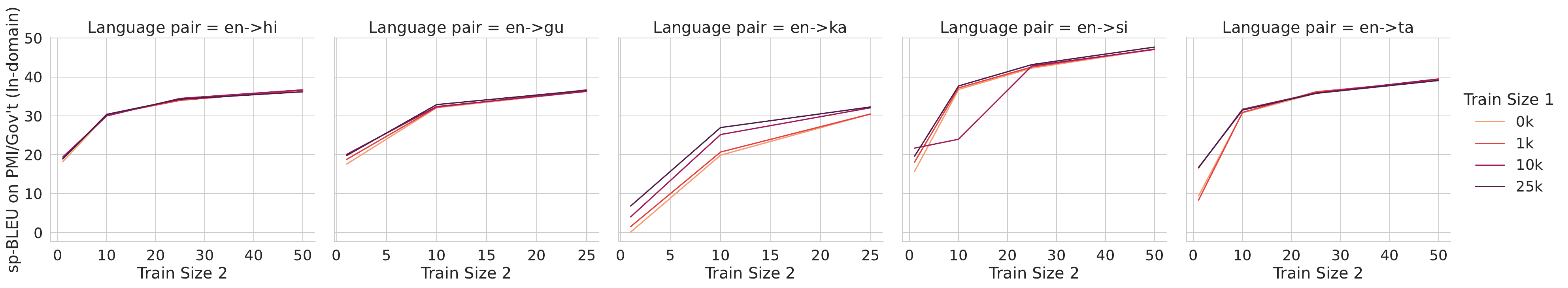}
        \caption{Intermediate Task Fine-Tuning with intermediate task - Bible and final task - PMI/Gvt. The spBLEU scores corresponds to scores on PMI/Gvt test set.}
     \end{subfigure}
     \begin{subfigure}[b]{ \textwidth}
         \centering
         \includegraphics[scale=0.22]{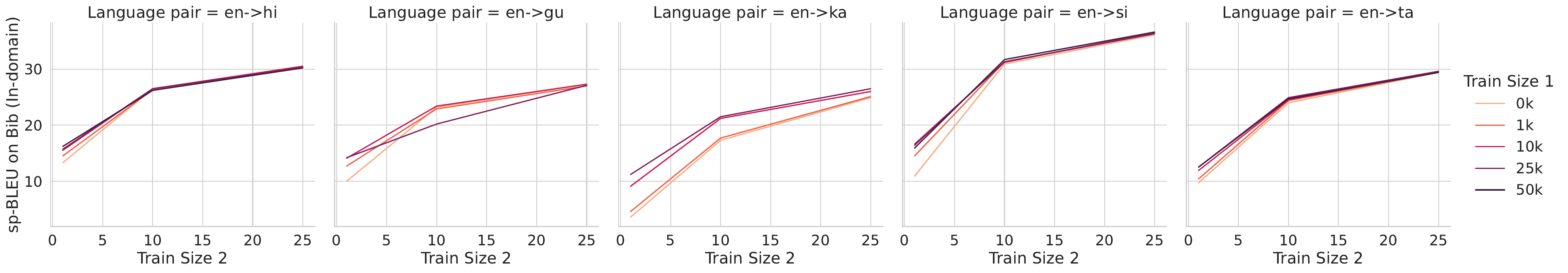}
        \caption{Intermediate Task Fine-Tuning with intermediate task - PMI/Gvt and final task - Bible. The spBLEU scores corresponds to scores on Bible test set}
     \end{subfigure}
     \caption{Intermediate Task Fine-tuning - effect of dataset sizes used for fine-tuning for in-domain case. The Y-axis represents the spBLEU scores and X-axis shows the final task size. Colours show the intermediate task size (size 0 = baseline). Train Size 1 corresponds to size of the intermediate task data set and similarly Train Size 2 for final task.}
     \label{fig:intermediate_size}
\end{figure}

\begin{figure}
     \centering
     \begin{subfigure}[b]{ \textwidth}
         \centering
         \includegraphics[scale=0.22]{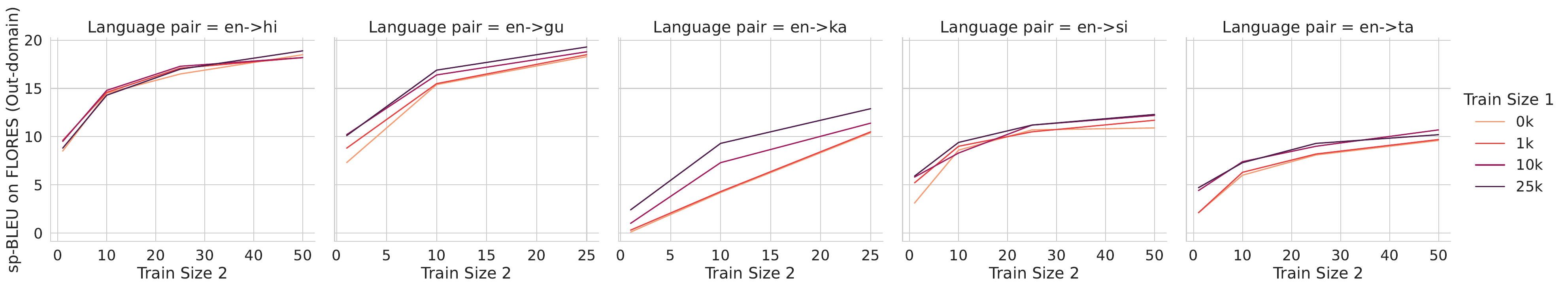}
        \caption{Intermediate Task Fine-Tuning with intermediate task - Bible and final task - PMI. The spBLEU scores corresponds to scores on FLORES test set.} %Train 1 Bib - Train 2 PMI
     \end{subfigure}
     \begin{subfigure}[b]{ \textwidth}
         \centering
         \includegraphics[scale=0.22]{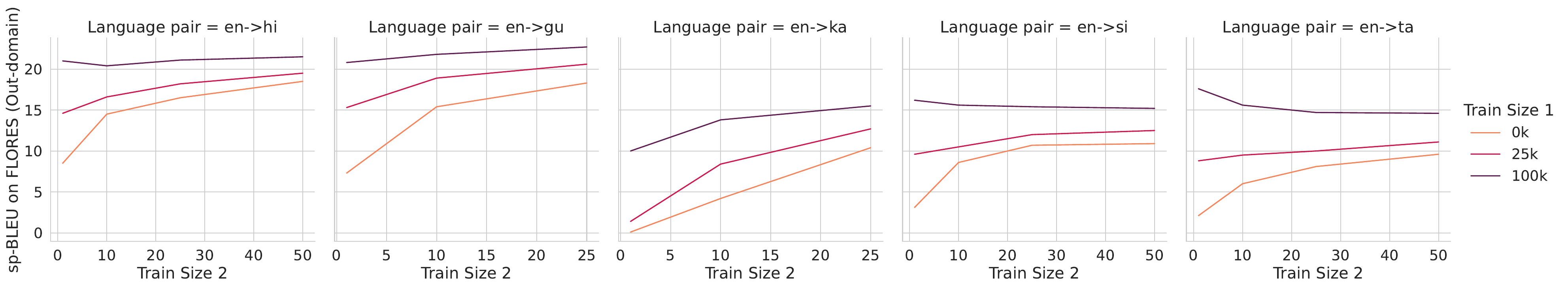}
        \caption{Intermediate Task Fine-Tuning with intermediate task - CC and final task - PMI. The spBLEU scores corresponds to scores on FLORES test set.} %Train 1 CC - Train 2 PMI
     \end{subfigure}
     \begin{subfigure}[b]{ \textwidth}
         \centering
         \includegraphics[scale=0.22]{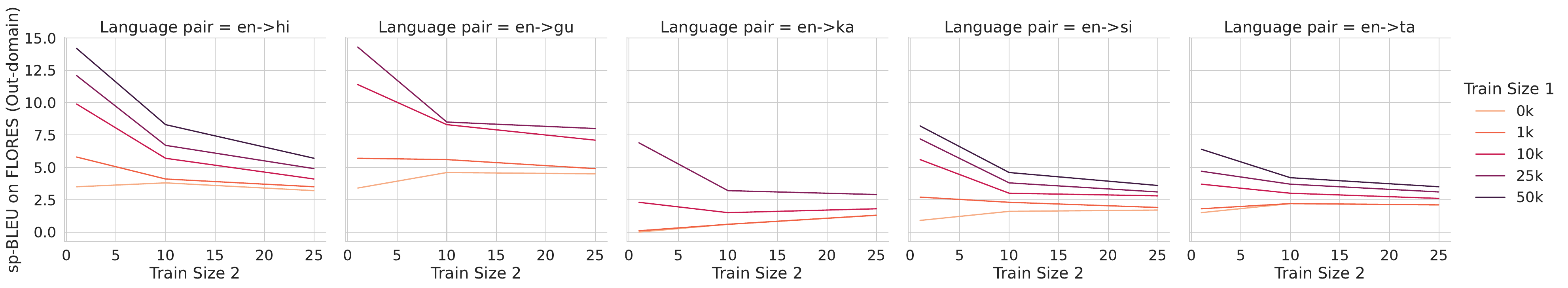}
        \caption{Intermediate Task Fine-Tuning with intermediate task - PMI and final task - Bible. The spBLEU scores corresponds to scores on FLORES test set.} %Train 1 PMI - Train 2 Bib
     \end{subfigure}
     \begin{subfigure}[b]{ \textwidth}
         \centering
         \includegraphics[scale=0.22]{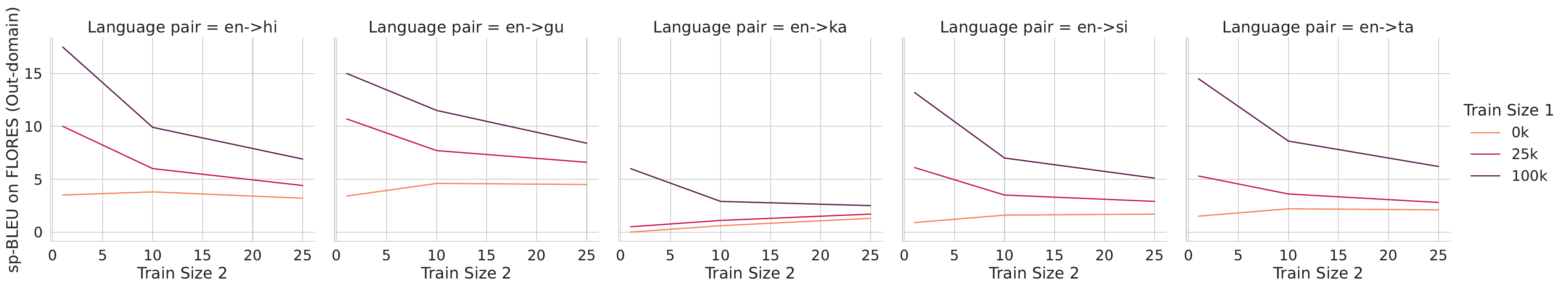}
        \caption{Intermediate Task Fine-Tuning with intermediate task - CC and final task - Bible. The spBLEU scores corresponds to scores on FLORES test set.} %Train 1 CC - Train 2 bib
     \end{subfigure}
     \caption{Intermediate Task Fine-tuning - effect of dataset sizes used for fine-tuning for out-domain case. The Y-axis represents the spBLEU scores and X-axis shows the final task size. Colours show the intermediate task size (size 0 = baseline). Train Size 1 corresponds to size of the intermediate task data set and similarly Train Size 2 for final task.}
     \label{fig:intermediate_size}
\end{figure}

\end{document}